\documentclass[conference]{IEEEtran}
\IEEEoverridecommandlockouts
\usepackage{cite}
\usepackage{amsmath,amssymb,amsfonts}
\usepackage{algorithmic}
\usepackage{graphicx}
\usepackage{textcomp}
\usepackage{xcolor}
\usepackage{subfigure}
\usepackage{comment}
\def\BibTeX{{\rm B\kern-.05em{\sc i\kern-.025em b}\kern-.08em
    T\kern-.1667em\lower.7ex\hbox{E}\kern-.125emX}}
\begin{document}

\title{Efficient Real-Time Image Recognition
Using Collaborative Swarm of UAVs and Convolutional Networks\\
}

\author{
    \IEEEauthorblockN{Marwan Dhuheir\IEEEauthorrefmark{1}, Emna Baccour\IEEEauthorrefmark{1}, Aiman Erbad\IEEEauthorrefmark{1}, Sinan Sabeeh\IEEEauthorrefmark{2}, Mounir Hamdi\IEEEauthorrefmark{1}}
    \IEEEauthorblockA{\IEEEauthorrefmark{1}Division of Information and Computing Technology, College of Science and Engineering,\\ Hamad Bin Khalifa University, Qatar Foundation, Doha, Qatar.
    }
    \IEEEauthorblockA{\IEEEauthorrefmark{2}Barzan Holdings QSTP LLC, Doha, Qatar.
   }
}

\maketitle

\begin{abstract}
Unmanned Aerial Vehicles (UAVs) have recently attracted significant attention due to their outstanding ability to be used in different sectors and serve in difficult and dangerous areas. Moreover, the advancements in computer vision and artificial intelligence have increased the use of UAVs in various applications and solutions, such as forest fires detection and borders monitoring. 
However, using deep neural networks (DNNs) with UAVs introduces several challenges of processing deeper networks and complex models, which  restricts their on-board computation. In this work, we present a strategy aiming at distributing inference requests to a swarm of resource-constrained UAVs that classifies captured images on-board and finds the minimum decision-making latency. We formulate the model as an optimization problem that minimizes the latency between acquiring images and making the final decisions. The formulated optimization solution is an NP-hard problem. Hence it is not adequate for online resource allocation. Therefore, we introduce an online heuristic solution, namely DistInference, to find the layers placement strategy that gives the best latency among the available UAVs. The proposed approach is general enough to be used for different low decision-latency applications as well as for all CNN types organized into pipeline of layers (e.g., VGG) or based on residual blocks (e.g., ResNet). 
\end{abstract}


\section{Introduction}

Over the last years, UAVs have been introduced as a better alternative to the traditional technologies in various applications such as monitoring and detecting objects from the sky \cite{kanistras2013survey}, rescue operations \cite{bejiga2017convolutional}, goods delivery \cite{sawadsitang2018joint}, etc.  UAVs have plenty of advantages as they are cost-effective, flexible to maneuver, and efficient for data collection. In addition, UAVs can cover a broad range of areas to detect events (e.g., forest fires) and access difficult regions that traditional monitoring tools cannot access, such as volcanoes. Recently, UAVs have been exploited to prevent the spread of COVID-19 by measuring the body temperatures or delivering the vaccines and supplies to areas that cannot be accessed with traditional tools \cite{wu20205g}. UAVs' exemplary performance motivated their use in more critical and complex missions such as monitoring borders and collecting data from battlefields. Some of these applications require a swarm of UAVs to improve the model's performance and achieve efficient results \cite{kim2018designing}. One typical application is to distribute the image classification tasks of a surveillance system into the swarm of UAVs in order to reduce the latency of predicting the final decision. 

The on-board classification is beneficial three-fold: Firstly, it decreases the dependency between the UAVs and the ground station; therefore, the on-board computation reduces the operational cost. Secondly, it reduces the latency of making the classification decision, i.e., the UAVs communicate with each other to receive the requests, make the final decision, and avoid remote transmission overhead. Third, knowing that UAVs frequently capture high-resolution images for classification even if incidents are rarely occurring and due to the harsh environments, affording stable bandwidth links to remote servers was problematic. However, on-board CNN inference raised several challenges that should be resolved. More specifically, the UAVs have restricted memory and computation operation units to address the arrived inference requests. This motivated us to adopt layers' distribution of online requests among UAV devices to benefit from resource sharing,  scalability, and network adaptability. 

Therefore, it is essential to find a model that suits the requirements of DNN into a resource-constrained swarm of UAVs. In this context, we propose a model that splits the DNN into layers and allocates each layer into one UAV. The connected UAVs share the output between them until reaching the final classification decision \cite{zhao2018deepthings}. By following this strategy, the classification decision will be decided locally. The current research focuses on studying the distribution of DNN models over resource-constrained devices without considering the real-time load of requests, memory, and computation restrictions. Hence, it is crucial to redesign a distribution DNN model that considers these constraints, which will be the contribution of this work. We propose a surveillance system model in our work, and we use CNN networks to classify the input image. This model receives the classification requests and finds the best UAV participants that collaborate to achieve the best latency while considering memory and computational resource constraints. The designed model is general enough to be applied to different CNN models organized into the pipeline of layers (e.g., VGG) or based on residual blocks (e.g., ResNet), which is not the case of previous works focusing only on sequential models \cite{disabato2019distributed,teerapittayanon2017distributed}.  

Our paper's novel contributions are presented as follows: (1) we introduce a system model that is composed of a swarm of UAVs capturing images and collaborating in real-time to classify these data using CNNs having different structures. (2) We formulate our joint approach as an optimization problem that seeks to minimize the latency of making the final classification decisions. This model considers the limitations of memory and computational units in the connected UAVs and the dynamic load of requests. (3) Due to the hardness of the proposed optimization problem, we introduce an online heuristic solution, namely DistInference, that relaxes the optimal solution to find the optimal placement of layers among different UAV participants and achieve the best latency of classification.

The paper is organized as follows: Section II presents the related work, and section III explains the system model. Section IV shows the performance evaluation of the optimal and heuristic solution. 

\section{RELATED WORK}

The distribution of CNN networks among IoT systems has been an exciting topic for many researchers. It has various scenarios and approaches, and we present some basic approaches in this section. Authors in \cite{disabato2019distributed} proposed to distribute inference requests on pervasive device units. The distribution is per layer in which the available devices present technological constraints. 
The authors focused on finding the optimal layer placement that reduces the latency of making the classification decision. However, they did not consider the online load of requests in their static distribution of CNN segments. Our work in this paper distributes different layers of each real-time incoming request into different UAVs in order to collaborate and classify the input image locally. 
The distribution of DNN has many advantages ranging from decreasing the classification latency to improving the system scalability and security. The authors in \cite{baccour2020distprivacy} explored distributing DNN over IoT devices in a surveillance system to minimize the latency of making the classification decision and improve the system privacy by introducing a model called DistPrivacy. They found that when the model is divided into small segments (i.e., feature maps) and distributed among a higher number of IoT devices, the data can be covered from untrusted devices, and the system privacy increases. 
However, our paper's work focuses on studying the distribution of DNN among multiple UAVs to minimize the classification latency and generalize the optimization problem to include different CNN architectures such as VGG, LeNet, ResNet, etc. This is not the case of the aforementioned works covering only typical DNN models organized into a pipeline of processing layers without any residual block. Because monitoring and detecting objects using a swarm of UAVs require high memory and computation resources, one solution was introduced is to divide the CNN models' layers between UAVs and Mobile Edge Computing (MEC) servers. The authors in \cite{yang2020offloading}, suggested that shallow layers of the CNN models are executed at UAVs. Meanwhile, higher layers are relayed to MEC servers. The intuition behind this strategy is to reduce the intermediate data size compared to the original data. At the same time, high computation capacities are not necessary to process the data in shallow layers. However, this approach increases the operational cost, and it is susceptible to latency and bandwidth status because of the continual transmission between MEC servers and UAVs. In our work, the classification tasks are done on UAVs while minimizing the classification latency and avoiding remote transmissions. At the same time, memory usage and computation tasks cannot exceed the predefined thresholds. 

Using UAVs as computing servers improves the time execution of the classification tasks; meanwhile, this scenario requires considering some parameters such as the minimum number of layers and requests that the UAV swarms need to start working. The authors in \cite{hu2018uav} explored using multiple UAVs as edge servers to do the computational tasks. The authors suppose that if all the available layers are fully employed in the swarm of UAVs, i.e., all UAV layers are busy executing tasks, the UAVs work as relaying devices and off-load the tasks into access points (APs). 
This study did not focus on distributing the layers of each request to UAVs; however, to decrease the latency, it is crucial to consider distributing the layers of each request. Our work uses a swarm of UAVs for calculating the computational tasks onboard UAVs, and different layers of each request are dispersed into the UAVs, and each UAV takes part in requests and processes it.
\begin{figure}[hbt!]
\centering
\includegraphics[scale = 0.5]{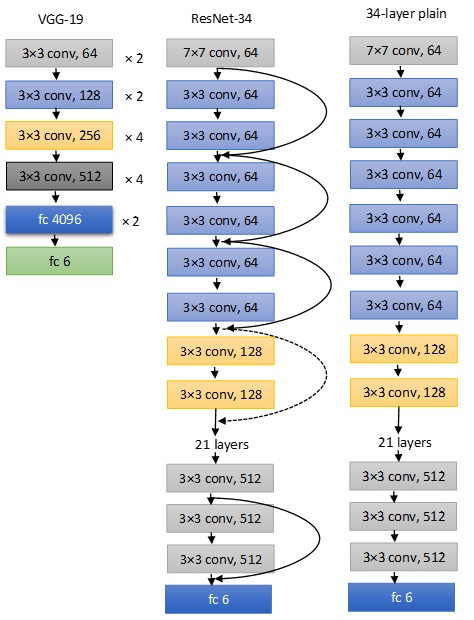}
\caption{different CNN architectures}
\label{fig:graph1}
\end{figure}

\section{DISTRIBUTED INFERENCES OVER A SWARM OF UAVS}
\subsection{System Model}

Our surveillance system model is shown in Figure 2. 
The model is composed of a set of UAVs that receives instructions to start the monitoring mission, such as forest fire detection, tracking a target on the ground, monitoring borders, etc. The captured images are distributed among the different UAVs, and each UAV executes a sub-task of the inference request. When defining the optimization problem, we consider the limitations of UAVs. These limitations are related to UAVs' available resources; therefore, we add the tolerated computation load and the maximum memory usage as constraints. In order to complete the classification in the UAVs' swarm, the UAV that does not meet the constrained resource limit will pass the request to the neighboring UAV to execute it. Our model uses a simplistic communication model as it is not the focus of this work, and each UAV uses a different transmission rate for transmitting and receiving data.
The scenario consists of a set of input sources represented by, $S=\{S_1,…,S_s\}$, capturing images to be classified by $N$ UAVs. Let us assume that the $i$ th UAV unit $i\in N_N$ is characterized by two important constraints, maximum memory usage $\bar{m_{i}}$, and maximum computational load $ \bar{c_i}$.

In this scenario, UAVs' swarm collaborates to execute the CNN model, i.e., the UAVs receive the input data and pass them to the CNN network models for classification. We assume that we have $r$ requests that belong to different CNN trained models. We also assume that $M_r$ is the number of layers representing the $r$-th number of requests in the considered CNN system. Accordingly, we have $j$ layers, for each $j \in\{1,…,M_r\}$, and $r \in N_r$, i.e., it represents the number of sub-tasks that is distributed among the available UAVs and required to classify the input image. The layer $j$ of the CNN model is characterized by two constraints: the memory usage $m_{r,k}$, and computational complexity $c_{r,k}$. The memory usage complexity $m_{r,k}$ (in bytes) is defined as the number of weights that the layer j can store multiplied by the size of the data type intended to characterize the parameters. The second requirement is the computational load $c_{r,k}$ that is defined as the number of multiplications that the system should execute as indicated in \cite{alippi2018moving}. Let us assume that $k_{r,j}$ be the memory occupation of the data transmitted from layer j to the next layer $j+1$ in the $r$-th network, $K_{r,s}$ be the memory occupation of the received image that is transmitted from the $r$-th source input to the unit executing the first layer of the $r$-th request in the CNN. Moreover, each UAV $ i \in \{1,…,N\}$ generates $R_i$ request in which $0 \le R_i \le R$.

Our proposed method includes residual-based architecture. This architecture adds advantages of training much deeper layers to decrease the latency of classification, improve classification accuracy, and improve the training error \cite{disabato2019distributed}. The residual block output receives two inputs, the one from the previous layer and the other from the shortcut connection layer. Therefore, we define $K_{r,j-\sigma}$  as the memory occupation of the layer’s output of the residual block from $j-\sigma$ to $j$ in the $r$-th request. The $\sigma$ here represents the number of strides in the residual block architecture. We also define the transmission rate $\rho_{i,k}$ that represents the transmission from the UAV $i$ to the UAV $k$, and the transmission rate of each UAV is different depending on the distance between UAVs, the quality of the link, and the available bandwidth.
\begin{figure}[hbt!]
    \centering
    \includegraphics[scale = 0.55]{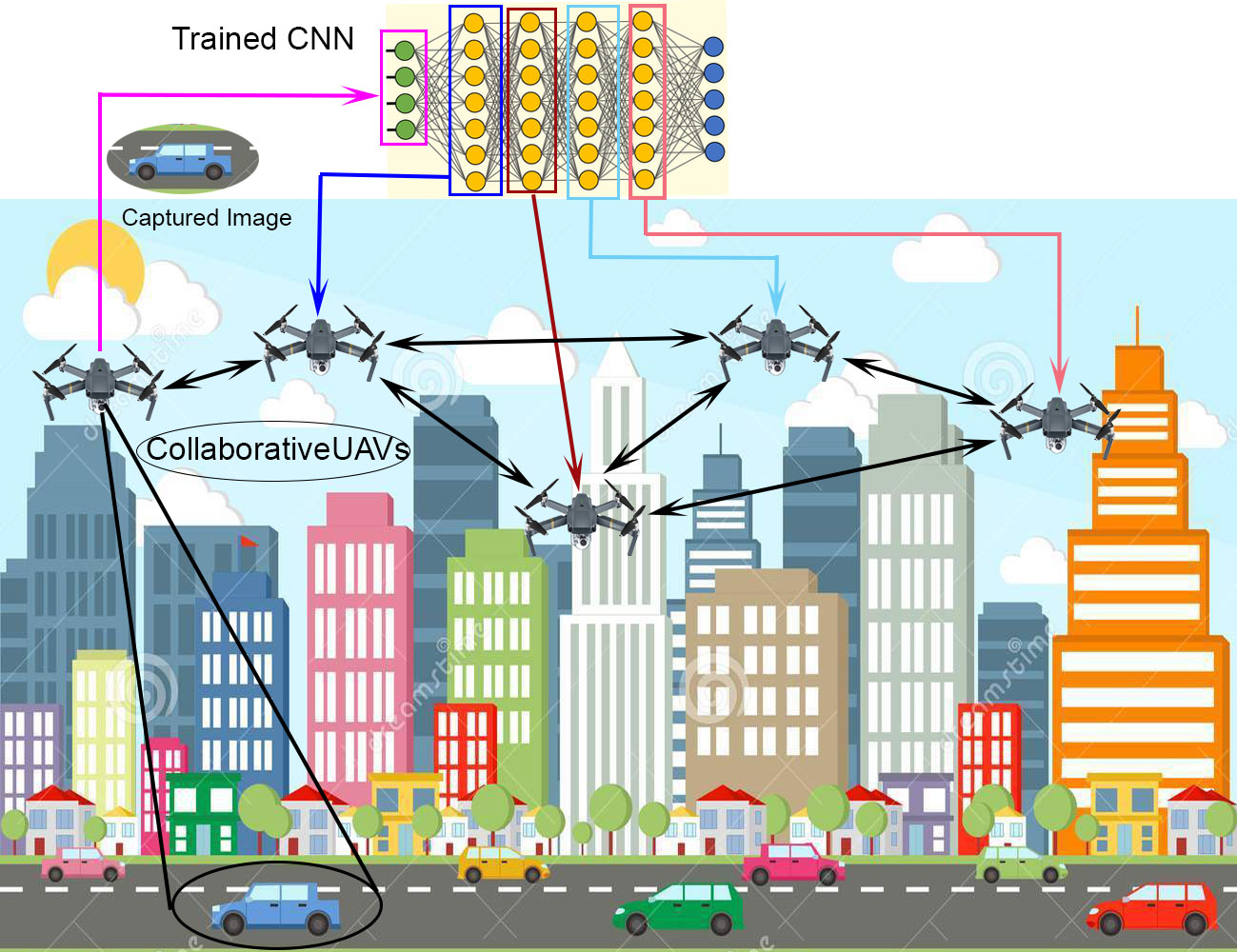}
    \caption{Surveillance System Model.}
    \label{fig:graph2}
\end{figure}

\subsection{Problem Formulation}

The optimization problem seeks to minimize the latency measured from capturing the images until making the decision. We formulate the objective function by calculating the sum of all arriving requests' transmission and processing time.

Therefore, the proposed optimization problem depends on two decision variables defined as follows:

$\forall r\in N_R,\forall i,k,n \in N_N, \forall j\in N_M$
    \begin{equation}
    \small
        \delta_{r,i,j} 
        = \left\{ 
        \begin{array}{l}
            1 \;\text{if UAV $i$ executes the $j$  layer of request $r$}\\
            0 \; \text{otherwise}\\
        \end{array}
        \right. 
    \end{equation}
   
    \begin{equation}
    \small
        \gamma_{r,n,k,j,\sigma} 
        = \left\{ 
        \begin{array}{l}
            1 \;\text{if the uav n transmits the output of the layer} \\ \text{$j-\sigma$ of request r to the uav k to process the} \\ \text{ layer j.}\\
            0 \; \text{otherwise}\\
        \end{array}
        \right. 
    \end{equation}

for $r\in N_R,i\in N_N$, and $j\in N_M=\{1,…,M\}$ and $M$ is the maximum number of layers that the request $r$ might have, i.e., $M = \max{(M_1,M_2,…,M_r)}$ which represents the maximum depth of the CNNs. The parameter $\sigma$ here represents the number of strides of the residual block in the CNN’s structure. The term $\gamma_{r,n,k,j,\sigma} \in\{0,1\}$ and it is the transmission of the output layers in the UAV that transmits from layer $j-\sigma$ to layer $j$. The requests to be distributed is $r$ requests, the two UAVs $n$ and $k$ represent the UAV that is outside the residual block and the one that is inside the residual block, respectively. Equation (1) presents the output transition between the two layers $j$ and $j+1$. In this scenario, we have two decision variables, $\gamma_{r,i,j,k,\sigma}$ and $\delta_{r,i,j}$. The objective function minimizes the latency by measuring the time of transmission from the layer $j$ to the next layer $j+1$ and measuring the transmission time from the residual layers. The latency is defined by the time between capturing  of size images $K_{r,s}$ by the source input $S_r$ and generating the corresponding decision after receiving the input of the last layer, sized $K_{r,Mr}$. The Integer Linear programming (ILP) optimization problem relies on the two decision variables $\gamma_{r,i,j,k,\sigma}$ and $\delta_{r,i,j}$ and the objective function of our problem is formulated as:

    \begin{equation}
    \begin{aligned}
    \small
        \min_{\gamma_{r,n,k,j,\sigma},
        \delta_{r,i,j}}
        ( \sum_{r=1}^{R}
        \sum_{i=1}^{N}
        \sum_{j=2}^{M}
        \sum_{k=1,k\neq i}^{N}
        \sum_{\sigma=1}^{j-1} 
        \max{(\gamma_{r,i,k,j,1}\;.} \frac{K_{r,j}}{\rho_{i,k}}\;,\\
        \gamma_{r,n,k,j,\sigma}\;.
        \frac{K_{r,j-\sigma}}
        {\rho_{i,k}}\;.
        \;\theta_{(j+1,j-\sigma)})}\;+
        \sum_{i=1}^{N} t_i ^{(p)} \;+
        t_s){
        \end{aligned}
    \end{equation}

s.t.
    \begin{equation}
    \small
        \sum_{r=1}^{R}
        \sum_{k=1}^{M} 
        \delta_{r,i,k} \;. \;m_{r,k} 
        \le 
        \bar{m_i}, \quad
        \forall i 
        \in{N_N}  
    \end{equation}

    \begin{equation}
    \small
        \sum_{r=1}^{R}
        \sum_{k=1}^{M} 
        \delta_{r,i,k} \;. \; c_{r,k} 
        \le 
        \bar{c_i}, \quad
        \forall i 
        \in{N_N}  
    \end{equation}

    \begin{equation}
    \small
        \sum_{i=1}^{N} 
        \delta_{r,i,j}  
        = \left\{ 
        \begin{array}{l}
            1 \;\text{\;if $j$ $ \le $ $M$}\\
            0 \; \text{\;otherwise}\\
        \end{array}
        \right. \quad
         \forall r 
        \in {N_R}, 
        \forall j 
        \in {N_M}, 
    \end{equation}    
    
    \begin{equation}
    \small
        \gamma_{r,i,k,j,\sigma}
        \le 
        \sum_{\sigma = 1}^{j-1} 
        \delta_{r,i,j-\sigma+2},\quad
        \forall r \in {N_R}, 
        \forall j \in {N_M}, 
        \forall {i,k} \in {N_N}, 
    \end{equation}
    
    \begin{equation}
    \small
        \gamma_{r,i,k,j} 
        \le 
        \delta_{r,k,j+1},
        \quad
         \forall r \in {N_R}, 
        \forall j \in {N_M}, 
        \forall {i,k} \in {N_N}, 
    \end{equation} 

    \begin{equation}
    \begin{aligned}
    \small
        \gamma_{r,i,k,j,\sigma} 
        \ge 
        \sum_{\sigma = 1}^{j-1}
        (\delta_{r,i,j-\sigma+2} \; + \; \delta_{r,k,j+1}\; - 
        \; 1), \\
         \forall r \in {N_R}, 
        \forall j \in {N_M}, 
        \forall {i,k} \in {N_N}, 
        \end{aligned}
    \end{equation}
    
and where 

    \begin{equation}
    \small
        t_s =
        \sum_{r=1}^{R}
        \sum_{i=1}^{N}
        \delta_{r,i,1} \;.
         \frac{K_{r,s}}{\rho_{s,i}}
    \end{equation}

    \begin{equation}
    \small
        t_i^{(p)} =
        \sum_{r=1}^{R}
        \sum_{i=1}^{N}
        \sum_{k=1}^{N}
        \delta_{r,i,k} \;.
         \frac{c_{r,s}}{e_i}
    \end{equation}   

Equation (3) presents the total latency of different inferences, which is composed of 4 parts:

\subsubsection{}
The source time $t_s$, defined in Equation (10). It is the time required to transmit the captured images from the source $S_r$ to the UAV unit computing the first layer.
\subsubsection{}
The processing time of the current working UAV unit $t_i^{(p)}$ that is explained in equation (11), and it is the time required for the total latency to compute all the layers assigned to the uav $i$. The processing time is defined as the time ratio between a computational load of $c_{r,k}$ that the layer needs to the number of multiplications $e_i$ that UAV $i$ can carry in a second.
\subsubsection{}
The transmission time between two devices $i$ and $k$, and it is defined as the time of transmitting the intermediate representation of the captured images of the UAV device $i$ to the UAV device $k$ in the swarm, and it is defined as,

    \begin{equation}
    \small
        \frac{K_{r,j}}{\rho_{i,k}}
    \end{equation}
where $\rho_{i,k}$ is the transmission data-rate from the $i$-th UAV unit device to the $k$-th UAV unit. The transmission rate $\rho_{i,k}$ represents the distance and the quality of the connection between UAVs. Since UAVs are moving, the value of $\rho_{i,k}$ is changing accordingly over interval of time, therefore, the optimization solution needs to re-run for each change of the network. Upon this point, we defer studying the impact movement of UAVs on classification for future work.
\subsubsection{}
The latency of transmitting the output of the residual layers which are represented by $\gamma_{r,n,k,j,\sigma}. \frac{K_{r,j-\sigma}}{\rho_{i,k}}. \theta_{(j+1,j-\sigma)}$. The latency from the parallel transmission is represented by $\max{(\gamma_{r,i,k,j,1}\;. \frac{K_{r,j}}{\rho_{i,k}}\;,\\ \gamma_{r,n,k,j,\sigma}\;. \frac{K_{r,j-\sigma}} {\rho_{i,k}}\;. \theta_{(j+1,j-\sigma)})}$, and it represents choosing either the transmission that comes from layer $j$ to the next layer$j+1$, or the transmission from layer $j-\sigma$ to the layer $j$

The parallel transmission in the residual block is composed of two paths. The first path is the transmission of the two successive layers, and the second path is the transmission between the successive layers and the shortcut connection that comes from the residual-based model i.e. the output from the residual block is the summation of the two paths. $\theta_{(j+1,j-\sigma)}$is zero if there is no residual blocks.

The constraints in equations (4) and (5) are set for the maximum memory usage, and the maximum tolerated computational load of the UAVs in the swarm, respectively. They are thresholds to the UAVs in which it confines the work to be within the allowable limit and to make the designed model more realistic. The constraint in equation (6) is set for making each layer is computed by only one UAV device.

The constraints in Equations (7), (8), and (9) are to ensure that the value of $\gamma_{r,i,k,j,\sigma}$ is equal to 0 if there is no transmission of the layer's output $j$ that is executed at the device $i$ and the next layer that is executed at the device $k$ and 1 if both of $\delta_{r,i,j-\sigma+2}$ and $\delta_{r,i,j+1}$ is equal to 1, i.e., it is 1 if both of them are 1 and zero if one of them is 0. To relax the optimization run-time and ensure the linearity of the problem, we added the second decision variable $\gamma_{r,i,k,j,\sigma}$, however, it could be replaced by $\delta_{r,i,j-\sigma+2}$ and $\delta_{r,k,j+1}$.

\subsection{Online Heuristic}

The optimization in Equation (3) is an NP-hard problem, which is not adequate for online resource allocation. Therefore, we propose an online heuristic solution that performs the real-time allocation. The proposed online greedy algorithm, namely DistInference, is explained in algorithm 1.  The process starts when an inference request is initiated, then CNN tasks are distributed among different UAVs to begin the classification process. The allocation of layers among the available UAVs is greedy, and the layers are assigned one by one on the swarm of UAVs. We choose the UAV that accomplishes the best latency with the maximum residual computation (line 8). The chosen UAV has to respect the memory and computation constraints. Consequently, the greedy allocation does not have any knowledge about the computation requirements; we choose the UAV that has the lowest $nrm(i) = \alpha \times t(j) + \beta / \bar{c_i}$ (line 11). From the function of $nrm$, a tradeoff is initiated between the latency and the residual computation of the UAV device. The chosen UAV will then be tested whether it respects the available resources of memory and computation $condi1$ (line 12); otherwise, it would be removed from the calculation. Following the calculations, any UAV that suffers from resource exhaustion would be rejected.

    \begin{figure}[hbt!]
        \centering
        \includegraphics[width=0.45 \textwidth]{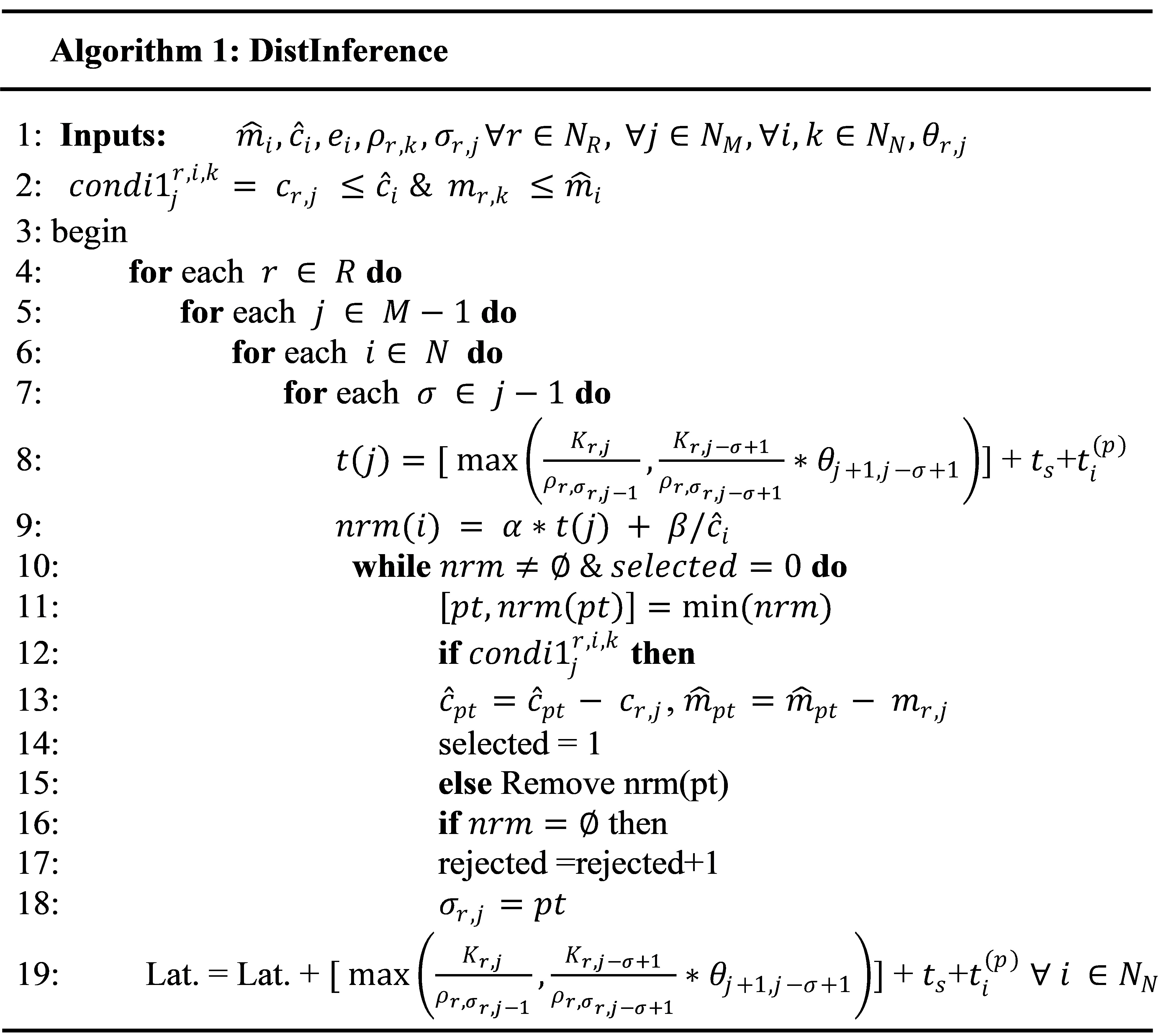}
        \label{fig:algorithm1}
    \end{figure}
    
\section{SIMULATION RESULTS AND EVALUATION}


This section illustrates the performance evaluation of our system model. We present the optimal solution results and compare the optimal solution with the heuristic solution. Furthermore, we present the result of the heuristic algorithm with different layers, requests, UAVs, and the minimum number of layers to start getting rejections. Moreover, we calculate the shared data and compare its result on ResNet and sequential layers model. This comparison is based on the number of requests and the length of CNN layers. We used Raspberry Pi 3B+ with a 1.4GHz 64-bit quad-core processor and 1GB of RAM in this simulation. The number of multiplications per second $e_i$ (defined as the number of 10ths of the clock cycle per second \cite{oh2019towards}) is $560 * 10^6$.

Figures 3(a) and 3(b) shows the results of the optimal latency with different requests and CNN layers, respectively. It is observed that as we increase the capacities of the system, the latency decreases accordingly. Figure 3(a) explains that as the request numbers increases, the latency increases, too. In this calculation, we used 5 CNN layers and 5 UAVs to receive the requests and distribute them. Figure 3(b) explains the relationship between increasing the length of the CNN layers and the latency of making the classification decision using 5 requests and 5 UAVs. It is shown that as the layer numbers increases, the latency rises accordingly. Figure 3(c) presents the minimum number of UAVs to start accepting requests and distributing them among the available layers to classify the captured input image.


\begin{figure}[h]
\centering
	\mbox{
	    \hspace{-7mm} \subfigure[\label{im1}]{\includegraphics[scale=0.23]{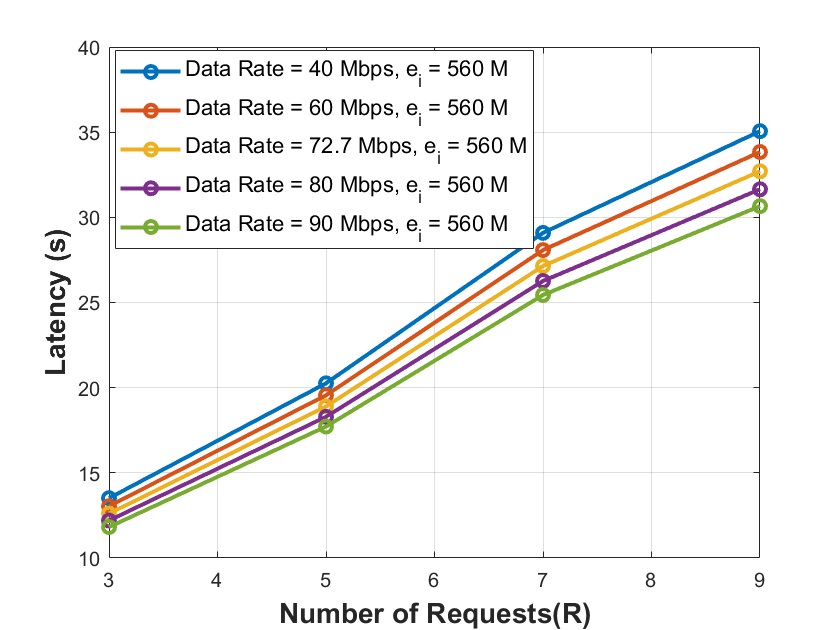}}
	     \hspace{-7mm}
	     \subfigure[\label{im2}]{\includegraphics[scale=0.23]{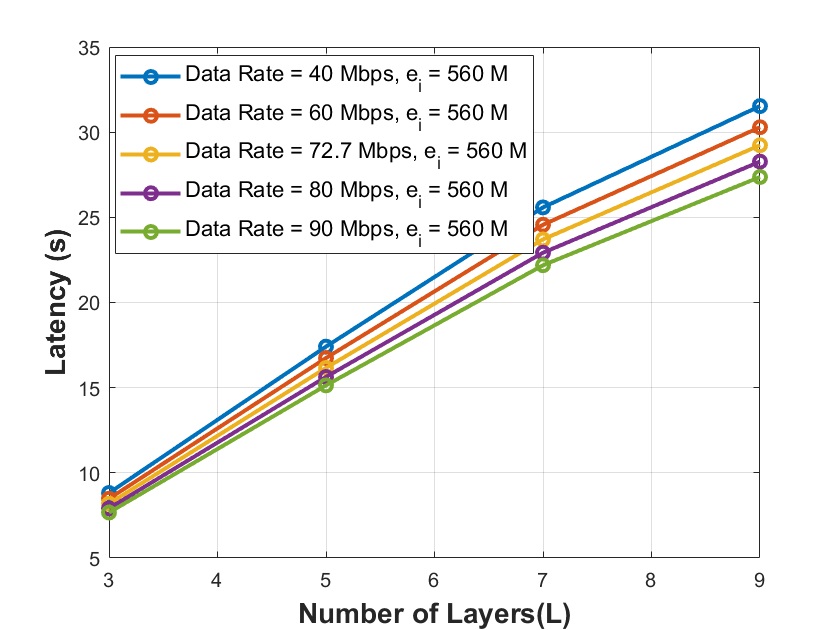}}
} 
	\mbox{
		\subfigure[\label{im3}]{\includegraphics[scale=0.23]{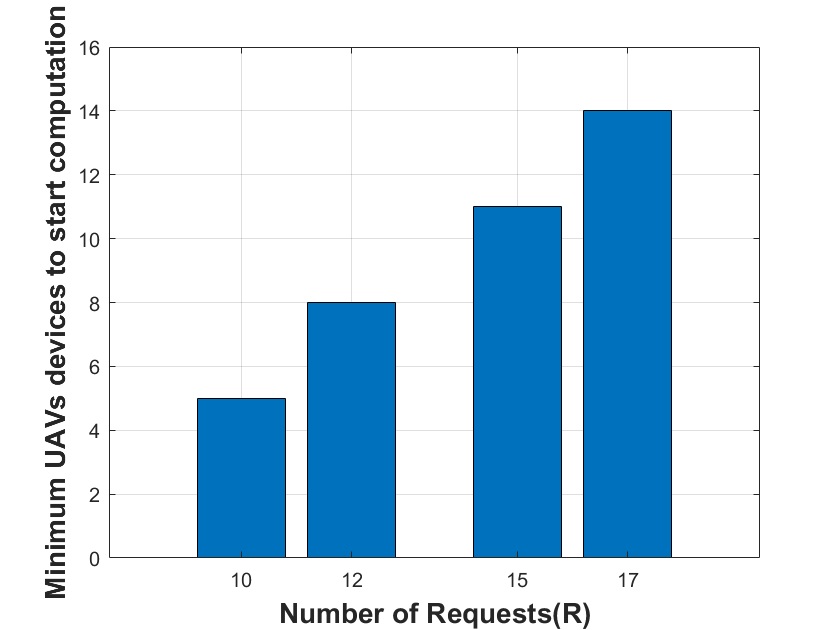}}
}
	\caption{Optimal solution simulation results.}
	\label{fig:graph3}
\end{figure}
   
Figure 4 illustrates the comparison between the optimal solution and heuristic solution with different $\alpha$s and $\beta$s. It is clear that the latency result of the optimal solution is better than that of the heuristic solutions. In this calculation, we used 5 CNN layers and 5 UAVs to receive the requests and distribute them. Moreover, the heuristic solution results with $\alpha$ and $\beta$ 0.7 and 0.3, respectively give the best solution among the others that is nearest to the optimal solution result. Therefore, we consider these values of $\alpha$ and $\beta$ in our calculations.
        \begin{figure}[hbt!]
            \centering
            \includegraphics[width=0.32 \textwidth]{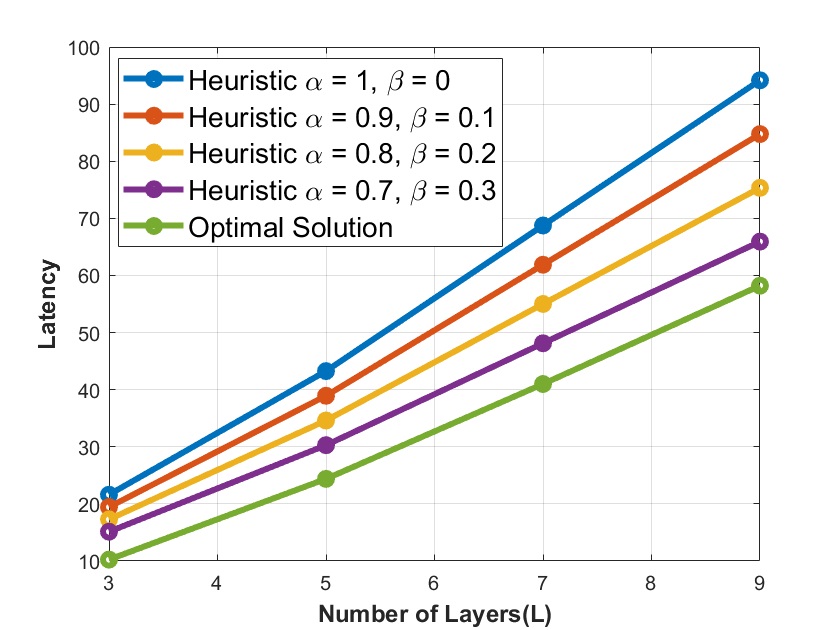}
            \caption{Optimal solution Vs heuristic with different $\alpha$ and $\beta$}
            \label{fig:graph5}
        \end{figure}
    
Figures 5(a), 5(b), and 5(c) illustrate the calculation of latency under the heuristic solution depicted in algorithm 1 with different requests, CNN layers, and UAV devices, respectively. It is observed that as we increase the capacities of the system, the latency decreases accordingly. Figure 5(a) explains that as the number of requests increases, the latency increases, too. Figure 5(b) illustrates the relationship between increasing the number of CNN layers and the latency. It is shown that as the number of CNN layers increases, the latency rises accordingly. To get these results, we used 70 requests and 30 UAVs. Figure 5(c) illustrates the latency results when we have different UAV devices in which we use ten requests and 10 CNN layers to handle the distribution of requests. It is clear that as we increase the number of UAVs, the latency decreases simultaneously. Moreover, when we add more UAV devices, the latency decreases simultaneously as more UAVs in the same area mean closer devices and better communication links. Figure 4(d) shows the maximum number of layers per request in which the requests start getting rejections. For example, with 10 requests and 30 UAVs, the maximum number of layers is 34, and above 34 layers, we get several rejections, and the UAV devices cannot handle all the requests correctly.

\begin{figure}[h]
\centering
	\mbox{
	    \hspace{-7mm} \subfigure[\label{im4}]{\includegraphics[scale=0.23]{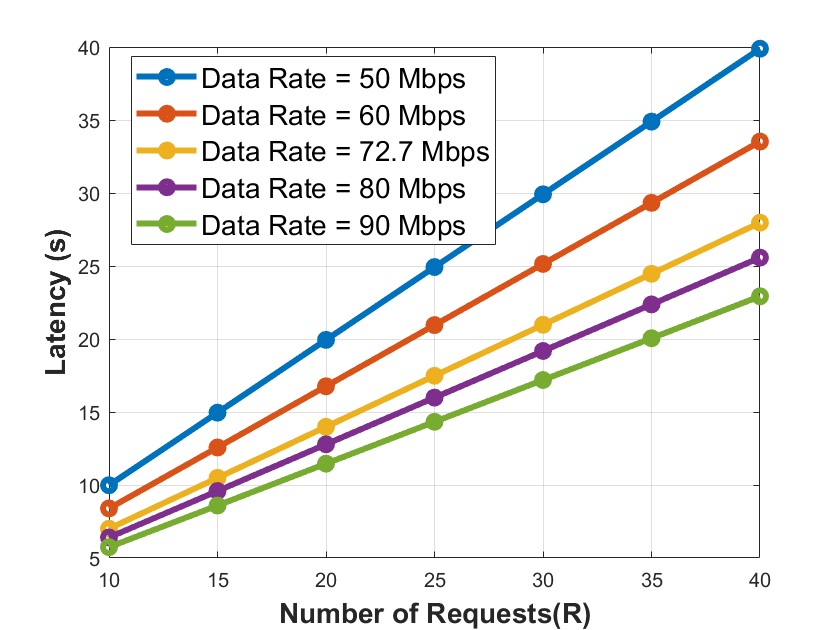}
	   }
	     \hspace{-7mm}
	     \subfigure[\label{im5}]{\includegraphics[scale=0.23]{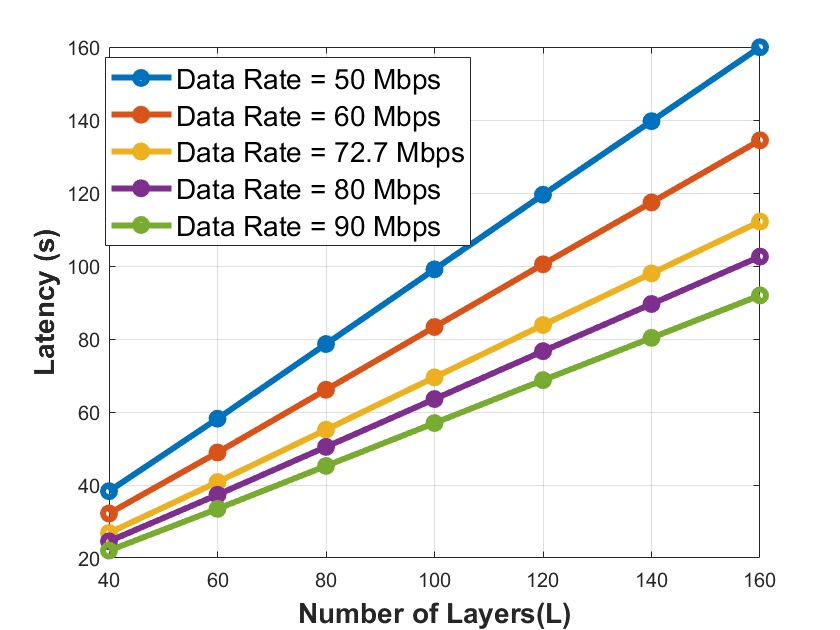}}
} 
	\mbox{
	\hspace{-7mm}
		\subfigure[\label{im6}]{\includegraphics[scale=0.23]{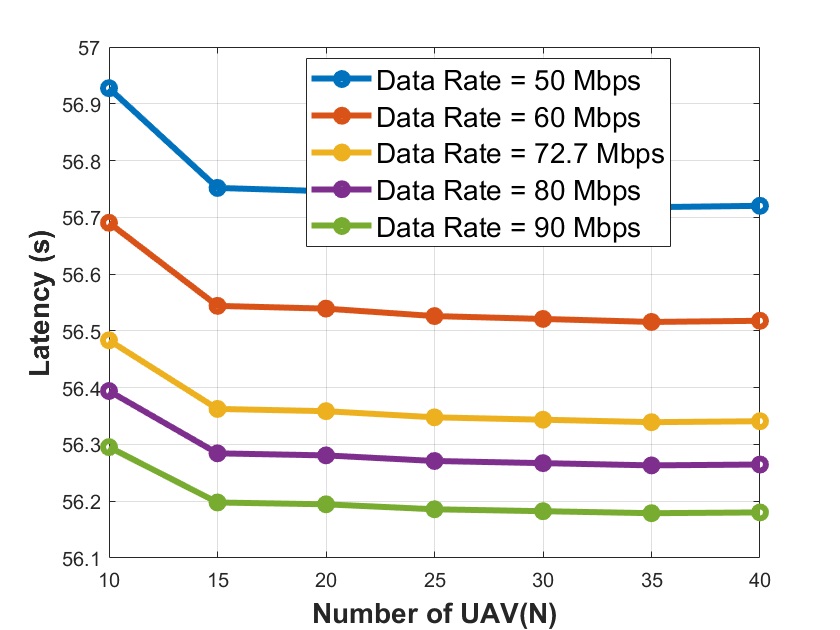}}
		\hspace{-7mm}
		\subfigure[\label{im7}]{\includegraphics[scale=0.23]{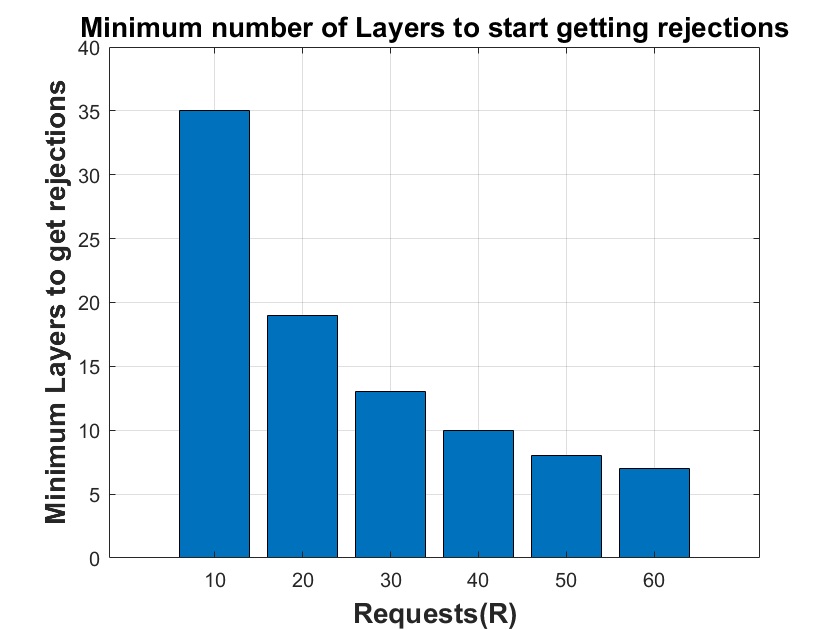}}
}
	\caption{Heuristic solution simulation results.}
	\label{fig:graph4}
\end{figure}
Figures 6(a) and 6(b) illustrate the calculations of shared data of heuristic solutions under different requests and CNN layers, respectively. It also compares sequential CNN models and ResNet under a different number of requests and layers. It is clear that sequential CNN models achieve less-shared data than ResNet as the sequential CNN models require less shared data and residual links need additional data to enter the layer.

\begin{figure}[h]
\centering
	\mbox{
	    \hspace{-7mm} \subfigure[\label{im8}]{\includegraphics[scale=0.23]{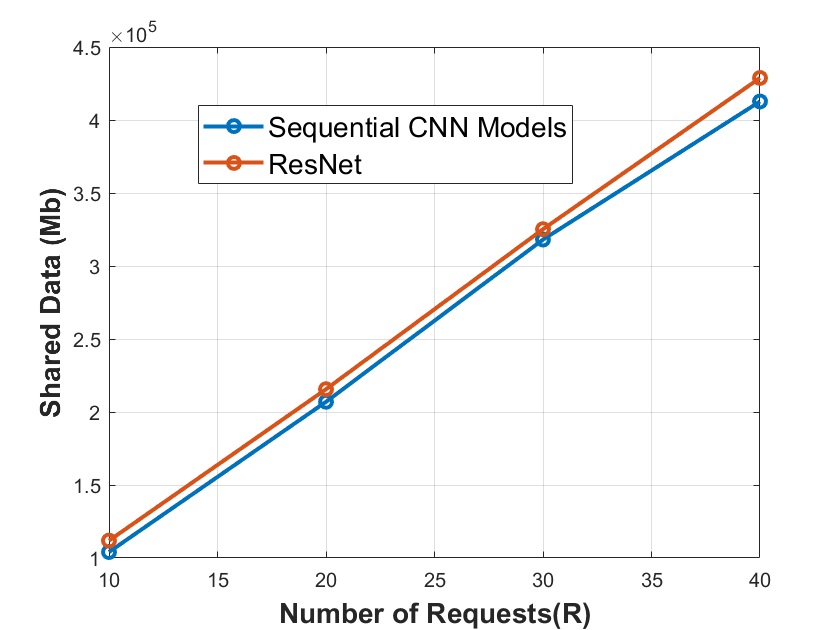}
	   }
	     \hspace{-7mm}
	     \subfigure[\label{im9}]{\includegraphics[scale=0.23]{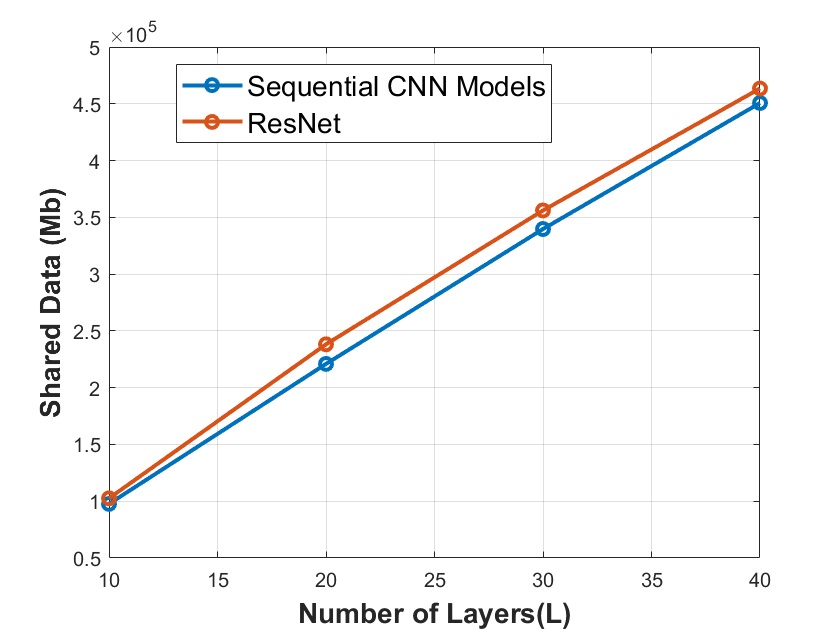}}
} 
	\caption{Shared data of ResNet and sequential CNN models.}
	\label{fig:graph6}
\end{figure}
\section{CONCLUSION}

In this paper, we presented the system model of distributing different layers of each request into a swarm of UAVs. Each UAV receives several requests and do part of the requests. The swarm of UAVs collaborates to find the minimum latency of making the classification decision. The distribution of the requests among the UAV swarm plays a significant role in reducing the classification latency. We formulate this as an optimization problem to calculate the minimum latency. The formulated optimization problem considers two main constraints: maximum memory usages and full computation complexity. Next, we introduced an online solution that is suitable for real-time scenarios. Our results proposed that as the requests and CNN models increase, the latency rises simultaneously. Our results also show that our proposed model gives less latency of making classification decisions than the online solution.

\section*{Acknowledgment}
This work was made possible by NPRP grant \# NPRP13S-0205-200265 from the Qatar National Research Fund (a member of Qatar Foundation). The findings achieved herein are solely the responsibility of the authors.  
\bibliographystyle{IEEEtran}
\bibliography{bibliography.bib}

\end{document}